\documentclass[3p,review,12pt]{elsarticle}

\usepackage{lineno,hyperref}
\modulolinenumbers[5]

\usepackage{siunitx}
\usepackage{graphicx}
\usepackage{multirow}

\graphicspath{ {images/} }

\begin{document}
\begin{frontmatter}

\title{MITOS-RCNN: A Novel Approach to Mitotic Figure Detection in Breast Cancer Histopathology Images using Region Based Convolutional Neural Networks}
    
\author[affdf8c4ec74725ec7dabd1cf591973d652]{Siddhant Rao\corref{cor1}\fnref{fn1}}
\ead{raosidd11@gmail.com}

\cortext[cor1]{Corresponding author}
\fntext[fn1]{Declarations of interest: none}
\address[affdf8c4ec74725ec7dabd1cf591973d652]{
    Monta Vista High School\unskip, 21840 McClellan Rd, Cupertino\unskip, CA}

\begin{abstract}
Studies estimate that there will be 266,120 new cases of invasive breast cancer and 40,920 breast cancer induced deaths in the year of 2018 alone. Despite the pervasiveness of this affliction, the current process to obtain an accurate breast cancer prognosis is tedious and time consuming, requiring a trained pathologist to manually examine histopathological images in order to identify the features that characterize various cancer severity levels. We propose MITOS-RCNN: a novel region based convolutional neural network (RCNN) geared for small object detection to accurately grade one of the three factors that characterize tumor belligerence described by the Nottingham Grading System: mitotic count. Other computational approaches to mitotic figure counting and detection do not demonstrate ample recall or precision to be clinically viable. Our models outperformed all previous participants in the ICPR 2012 challenge, the AMIDA 2013 challenge and the MITOS-ATYPIA-14 challenge along with recently published works. Our model achieved an F-measure score of 0.955, a 6.11\% improvement in  accuracy from the most accurate of the previously proposed models.
\end{abstract}
\end{frontmatter}
\section{Introduction}

One in eight U.S. women will develop invasive breast cancer at some point in their lives, placing breast cancer as the second most commonly diagnosed form of cancer, regardless of gender \citep{bcancer}. The World Health Organization recommends the use of the Nottingham Grading System for tumor grading \citep{elston}. The Nottingham Grading System is derived from the assessment of three main morphological features: nuclear atypia, mitotic count and tubule formation. Nuclear atypia is described as the deformation of nuclei in a population of cells and is characterized by the following factors: size of nuclei, size of nucleoli, density of chromatin, thickness of nuclear membrane, regularity of nuclear contour, and anisonucleosis (size variation within a population of nuclei). Tubule formation is described as the percent of cancer cells that are in regular tubule formation. As the cancer becomes more belligerent, the tumor cells proliferate via mitosis (the process of cellular division), making the mitotic count of a tumor an important prognostic factor. For this study, we will be focusing on the most documented and salient feature involved in an accurate breast cancer prognosis: mitotic count. Mitotic count needs little to no professional interpretation, due to the simple metrics used to identify proliferation rates using the mitotic count per high power field (HPF's: the area visible under the maximum magnification power of a microscope) : 0-9 mitoses per 10 HPF's is low proliferation, 10-19 mitoses per 10 HPF's is moderate proliferation and more than 19 mitoses per 10 HPF's is severe proliferation.

Despite the prevalence of breast cancer, current methods for breast cancer prognosis are quite primitive. Trained pathologists are needed to examine hundreds of high power fields of histology images. Biopsies often take around two to ten days for results to return to the patient \citep{bcancer}. Given the growing number of breast cancer incidences \citep{bcancer}, the traditional method for breast cancer prognosis is not sustainable. A computational approach would be a much more time and cost effective alternative, allowing for a streamlined breast cancer prognosis pipeline. This would allow for the deploying of pathological services to impoverished areas and the optimization of care centers globally.

\begin{figure}[h]
\centering
\includegraphics[height=0.11\textheight, width=0.2\textwidth]{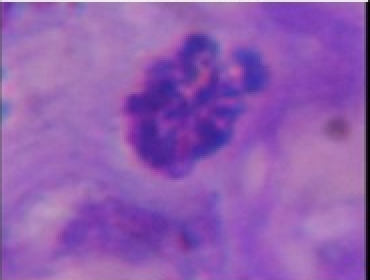}
\includegraphics[height=0.11\textheight, width=0.2\textwidth]{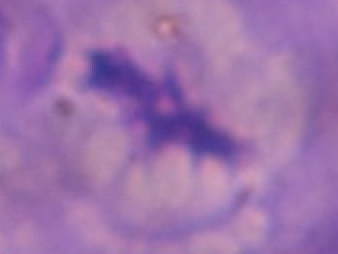}
\includegraphics[height=0.11\textheight, width=0.2\textwidth]{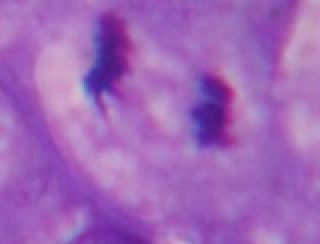}
\includegraphics[height=0.11\textheight, width=0.2\textwidth]{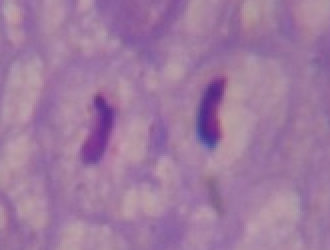}
\caption{(From left to right) The prophase, metaphase, anaphase and telophases stages of mitosis. The shape variation across stages can cause low mitotic figure detection accuracy. }
\label{fig:phases}
\end{figure}

However, there are some complications that limit the accuracy of both computational and manual mitotic count extraction. Obtaining an accurate mitotic count is quite a challenge, as mitoses are often of low density, but high variation throughout HPF's \citep{mitosis-atypia-14}. Such variation is seen across the four phases of mitosis (prophase, metaphase, anaphase and telophase), with each phase having its own distinct size and shape (see Fig. \ref{fig:phases}). Mitotic figures in the anaphase or telophase stage of mitosis are often misclassified as 2 mitotic figures rather than 1. The low density of mitotic figures is evident in the metrics used to classify various cancer severity levels using mitotic counts: 0-9 mitoses per 10 HPF's is low proliferation, 10-19 mitoses per 10 HPF's is moderate proliferation and more than 19 mitoses per 10 HPF's is severe proliferation. On average there are about 0-2 mitotic figures per HPF. Low density and high variation of mitotic figures makes scanning through hundreds of HPF's a tedious task when done manually and makes the practice susceptible to human error. For example, apoptotic cells (cells undergoing preprogrammed cell death) and other debris accumulated while preparing the tissue sample are often confused with mitoses due to their shaded, circular appearance. Irregularities in hematoxylin and eosin (H\&E) staining across cancer research/treatment centers also add to the variation of breast cancer histopathology images. 

Prior computational approaches to mitotic figure detection in breast cancer histopathological images within the scope of contests do not generalize well to new sets of data, resulting in relatively poor performance on an evaluation dataset. Outside the scope of participants in a mitotic figure detection challenge, several improvements have been made, but the methods are not accurate enough for clinical viability.

Deep learning is a growing field geared towards multi-scale pattern detection using deep neural network architectures. Adaptations of models like the convolutional neural network (CNN) can extract high level features from images to be used for object detection tasks like obtaining a mitotic count. One example of such a model is the Faster-RCNN proposed by \citep{NIPS2015_5638}, which uses features from an image to produce spatial coordinates for bounding boxes associated with certain classes.

We propose MITOS-RCNN: an adaptation of the Faster-RCNN model geared towards the automatic detection and counting of mitotic figures in breast cancer histopathology images. Our model was trained using the ICPR 2012, AMIDA 2013 and MITOS-ATYPIA-14 challenge datasets. We later compare the results of our models when fed sample images to those of previous works and demonstrate that our model significantly outperforms all other approaches.

\section{Related Work}

\subsection{Deep Learning}

Although deep learning methodologies have just recently begun to gain popularity, the underlying theory and applications have been present for quite some time. One of the earlier applications of deep learning for image analysis was the work done by \cite{6795724} using CNN's for handwritten zip code classification. However, support vector machines outperformed CNN's during this time due to the lack of computational resources available for deep learning methodologies to be successful. \cite{Krizhevsky:2012:ICD:2999134.2999257} improved upon the work done by \cite{6795724} with the introduction of CNN's for general object image classification and outperformed all existing methods in the ImageNet Large Scale Visual Recognition Challenge, thus showing the promise of deep learning techniques. \cite{2013arXiv1311.2524G} unified object classification tasks and object detection tasks with the original RCNN model. Models like the Fast-RCNN \citep{DBLP:journals/corr/Girshick15} and the Faster-RCNN \citep{NIPS2015_5638} improved the speed at which these models would be trained and evaluated, resulting in close to real-time object detection. 

\subsection{Mitotic Figure Detection}

Prior contests have been held with the sole purpose of discovering novel approaches to detecting mitotic figures in histology images such as MITOS challenge at ICPR 2012 \citep{roux} and the AMIDA 2013 challenge \citep{d00fb7c37eb248e1a28dc416fff2f8c3}. The winners of MITOS contest, \cite{10.1007/978-3-642-40763-5_51}, utilized a deep, max-pooling CNN which operates on patches of pixels and their respective color channels and classifies those pixels as mitotic or not mitotic figures, ultimately achieving an F-measure score of 0.782. The model was trained to formulate features based on training images, contrasting the other contestants use of handcrafted features. Ciresan et al.'s work was one of the first applications of CNN's in a histopathological context.

The AMIDA 2013 challenge \citep{d00fb7c37eb248e1a28dc416fff2f8c3} proved to be quite similar, as contestants either employed classifiers (e.g. random forest classifiers) that relied upon hand-crafted features or utilized deep learning methodologies similar to those previously proposed in the MITOS challenge. Ciresan et al. \cite{10.1007/978-3-642-40763-5_51} prevailed once again with the use of Multi Column Max-Pooling Convolutional Neural Networks (MCMPCNN). This new approach utilized a probabilistic representation of whether a pixel was a mitosis or not along with three 10-layer networks working in tandem. \cite{10.1007/978-3-642-40763-5_51} achieved an F-measure score of 0.611 with this approach.

Winners of the MITOS-ATYPIA-14 challenge achieved an F-measure score of 0.356 using a model called the Deep Cascaded Network \citep{AAAI1611788} consisting of the following main steps: (1) candidate mitotic figure detection using a fully convolutional network and (2) discrimination classification of the detected mitotic figures candidates using a CNN. 

Recent advances in mitotic figure detection outside the scope of contests have been made by various works. For example, \cite{7405343} employed a CNN paired with a crowd-sourced learning architecture and achieved an F-measure of 0.433. Additionally, Saha et al. \cite{saha} use both hand-crafted and learned features in their proposed model, achieving an impressive F-measure score of 0.900.

Our proposed approach builds upon the most successful prior works by utilizing a modified Faster-RCNN tuned for the detection of small objects, which matches the speed of previous CNN implementations but with more accurate detections.

\section{Materials}

\begin{table}
 \begin{tabular}{||c c c c||} 
 \hline
 Scanner & Dimensions of x40 frame (px) & Res. at x40 (\SI{}{\micro\meter}/px) & \# of Frames    \\
[0.5ex] 
 \hline\hline
 Aperio & 1539 * 1376 & 0.2455 & 2622 \\ 
 \hline
 Hamamatsu  & 1663 * 1485 & 0.2273   & 2016 \\ 
 \hline
\end{tabular}
\caption{The dimensions, resolution and number of frames from each of the two scanners: the Aperio Scanscope XT and the Hamamatsu Nanozoomer 2.0-HT. }
\label{table:data}
\end{table}

\subsection{Dataset Description}

Our evaluation set of data consists of 100 samples of each of the ``mitotic figure" and ``not mitotic figure" classes spread across 187 HPF subsections.

\section{Methods}

\subsection{Data Preprocessing and Augmentation}

Our model takes in only 299x299 px images as input and resizes any input to this dimension. We split each image into 16 equal subsections to make sure minimal downsampling of each image takes place. This allows for more ``attention" to small-scale features. 
Since the Faster-RCNN outputs bounding box spatial coordinates, we introduced a new dataset, MITOS-BOXES, using the preexisting centroid coordinates and annotating box regions by hand for each mitotic and non-mitotic figure.
Due to the relatively small size of our dataset, we introduced artificially augmented versions of existing data to create more samples for the model to learn from and to prevent overfitting. We rotated all images by preset values (90$^{\circ}$, 180$^{\circ}$, 270$^{\circ}$), as we believed that keeping raw pixel data intact  would be beneficial for the model. Due to the inconsistency of staining techniques across the data sources, all image data was normalized via a procedure described by \cite{e426b70b6aef4f4cba21511905c8236a} (original samples were kept in dataset). The final size of the dataset was 37,104 HPF's. 

\begin{figure}[h]
\centering
\includegraphics[width=\textwidth]{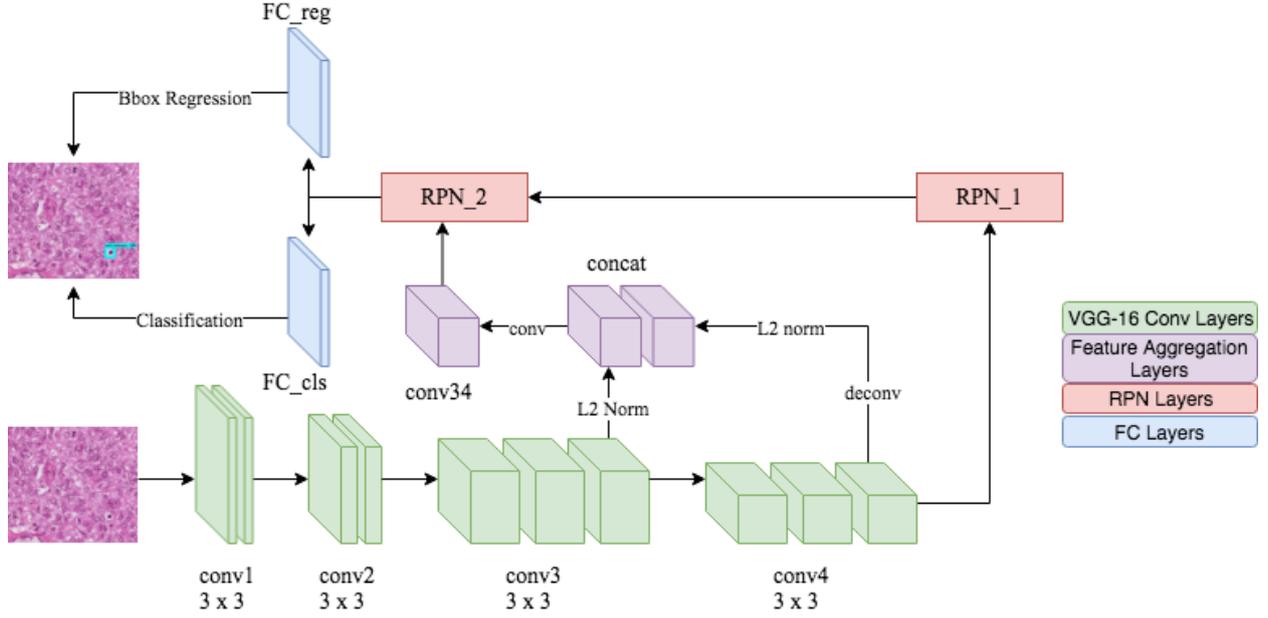}
\caption{The architecture of our proposed MITOS-RCNN model, from the VGG-16 convolutional layers to the fully-connected layers outputting the final detections.}
\label{fig:diagram}
\end{figure}

\subsection{Proposed Model}

Our proposed model (see Fig. \ref{fig:diagram} for diagram of architecture) is a variation of \cite{NIPS2015_5638} proposed Faster-RCNN model and is composed of two main modules: (1) a region proposal network (RPN) which returns regions of interest (ROI's) and (2) a detection network which classifies and discriminates regions of interests while performing a bounding box regression, ultimately returning spatial coordinates with associated classes. We utilized the VGG-16 model \citep{simonyan} as our base feed-forward CNN in order to extract powerful hierarchical features from an input image. We refer to the 'conv1\_3', 'conv2\_3', 'conv3\_3', 'conv4\_3' and 'conv5\_3' layers of the VGG-16 network as $conv_1$, $conv_2$, $conv_3$, $conv_4$, $conv_5$, respectively. Region proposals are generated during the two-stage top-down cascade multi-scale proposal generation process, where features from both the $conv_3$ and $conv_4$ feature maps are aggregated and used by two separate RPN's: $RPN_1$ and $RPN_2$. The generated region proposals are then fed into two sibling fully connected layers ($FC_{reg}$ and $FC_{cls}$) to regress bounding box spatial coordinates and classify the generated region proposals as being either ``mitotic figure" or ``not mitotic figure".

\subsubsection{Small Scale Object Detection}

\cite{8019550} proved that with the standard Faster-RCNN architecture, the minimum detectable object size was around 44px. This is due to the loss of information as the feature map representation dimensionality is reduced via the pooling layers within the feed-forward network. Essentially, the later $conv$ layer feature maps contain rich abstract-level features which are far too coarse to be used for extraction of features pertaining to small objects. A 44px object detection threshold is not optimal for the detection of mitoses, as the average size of the mitotic figures in our MITOS-BOXES dataset was around 30px. To avoid this issue we omit the $conv_5$ layer and utilize the feature maps from only $conv_3$ and $conv_4$, allowing for the minimum detectable object size to be 15px and 22px, respectively \citep{8019550}.

\subsubsection{Two-Stage Top-Down Cascade Multi-Scale Proposal Generation}

Our adapted RCNN model generates multi-scale proposals at two distinct stages of the network using features aggregated from multiple convolutional layers in a top-down manner in order to obtain more refined proposals for specifically small-scale objects. This allows the model to use semantic knowledge from both higher and lower level features to produce accurate object detections. $RPN_1$ follows the $conv_4$ layer and generates around 15k proposals (most of which are discriminated via NMS thresholding of 0.7). $RPN_2$ utilizes features aggregated from a feature map consisting of both $conv_3$ and $conv_4$ concatenated. To obtain a concatenated feature map, the $conv_4$ feature map is upsampled by a subsequent deconvolutional layer in order to obtain a resolution matching that of $conv_3$. Then, we normalize each layer using L2 norm, concatenate the two feature maps to obtain $conv_{3 + 4}$ and reduce the $conv_{3 + 4}$ feature map to a dimension of 256x1x1. $RPN_2$ utilizes inputs from two sources: (1) the proposals from $RPN_1$ and (2) the output of a ``sliding window" operating on the $conv_{3 + 4}$ feature map with a scale of $64^2$px and a 1:1 aspect ratio. Proposals are further refined by the model by discriminating low-quality proposals generated by $RPN_2$ via NMS thresholding of 0.7 and then fed into an ROI pooling layer to normalize region proposal scales.

\subsubsection{Detection Network}

The detection network of our model is identical to that of the RCNN's proposed by \cite{NIPS2015_5638} and \cite{DBLP:journals/corr/Girshick15}. Two sibling output layers make up this detection network: a bounding box classification layer and a bounding box regression layer. Both networks are fully-connected layers and receive the refined region proposals from the $RPN_2$ layer of the model after the proposal scales have been normalized by the ROI pooling layer. The bounding box classification layer, $FC_{cls}$, outputs a softmax probability distribution, $p = (p_0 ... p_k)$, over the $0$th $... k$th classes for every region proposal, where the softmax function is defined as: 
\begin{equation}
P(c_i|x) = \frac{\exp(y_i)}{\sum^{N}_j \exp(y_j)}.
\end{equation}
Given an input, $x$, the softmax function computes the probability of a class, $c_i$,  using the classification score for the $i$th class, $y_i$, and the classification scores for all the classes.

The fully connected layer used for bounding box regression, $FC_{reg}$, outputs a vector of regression offsets, $t^k = (t^k_x, t^k_y, t^k_w, t^k_h)$, specifying a scale-invariant transformation (of the top-left corner of the box $(x,y)$ and the width and height of the box $(w,h)$) to the input region proposal coordinates for each of the $k$ classes.

\subsubsection{Training}

The loss function being minimized during the training process takes into account the loss from both modules of the detection network - the regression network, $L_{reg}$, and the classification network, $L_{cls}$ - and is defined as:
\begin{equation}
L({p_i, t_i})= \frac{1}{N_{cls}} \sum_i L_{cls}(p_i, p_i^*) + \lambda \frac{1}{N_{reg}} \sum_i p_i^* L_{reg}(t_i, t_i^*),
\end{equation}
where $i$ is the anchor index (anchors are synonymous with  bounding box proposals), $p_i$ is the ``objectness" of the anchor $i$, $p_i^*$ is the predicted ``objectness" of the anchor $i$, $t_i$ is the coordinate vector of the bounding box prediction, $t_i^*$ is the coordinate vector for the ground truth bounding box with a positive anchor, $N_{cls}$ is the batch size, $N_{reg}$ is the number of anchors, $L_{cls}$ is the log loss over the object and not object classes,  $L_{reg} = R(t_i - t_i^*)$ ($R$ is the robust loss function described by \cite{DBLP:journals/corr/Girshick15}) and $\lambda$ is a balancing hyperparameter (set to 10 in our implementation).

Our model was trained in the same fashion as the Faster-RCNN \citep{NIPS2015_5638}. Weights in all layers of the RCNN were initialized from a zero-mean Gaussian distribution with standard deviation of 0.01, while all VGG-16 layers remained with their pre-trained weights. We trained the fine-tuned the RCNN model with our proposed MITOS-BOXES dataset using standard stochastic gradient descent (SGD) \citep{6795724}. Batch size was kept to 10 images. The model was trained for 60,000 mini-batches with a learning rate of 0.001 and then 20,000 mini-batches with a learning rate of 0.0001. A momentum of 0.9 and a weight decay of 0.0005 was used.

\subsection{Implementation Details}

Both the training and testing process were performed on the Google Cloud Platform ML Engine. We used 5 NVIDIA Tesla K80 GPUs and 3 parameter servers in order to distribute the training process. The testing process only required a single NVIDIA Tesla K80 GPU.  Dataset augmentation/preprocessing and model evaluation was done locally on a computer running the MacOS High Sierra operating system with a 2.7 GHz Intel Core i5 processor and 8 GB of RAM. Our implementation was in the Tensorflow machine learning framework \citep{45381}.

\section{Results}

\begin{figure}[h]
\centering
\includegraphics[height=0.22\textheight, width=0.4\textwidth]{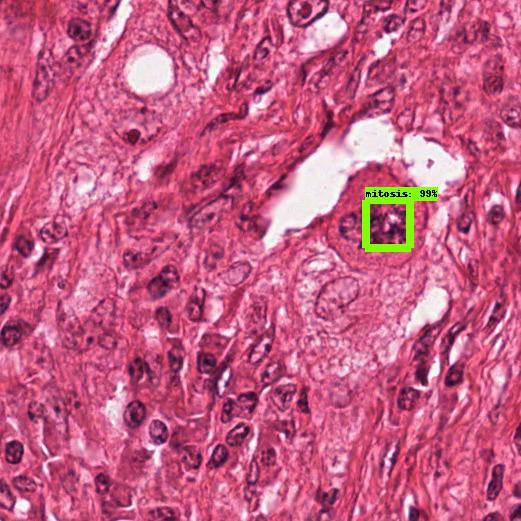}
\includegraphics[height=0.22\textheight, width=0.4\textwidth]{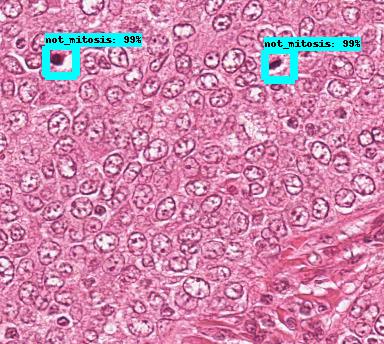}
\includegraphics[height=0.22\textheight, width=0.4\textwidth]{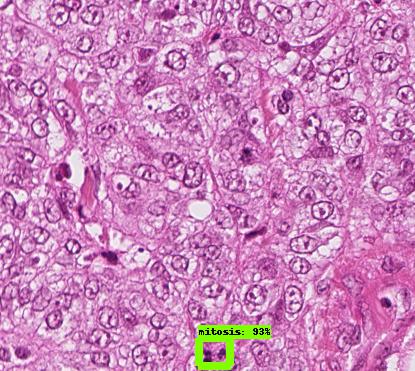}
\includegraphics[height=0.22\textheight, width=0.4\textwidth]{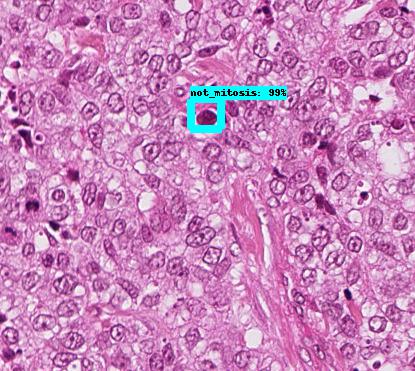}
\caption{Sample detections of ``mitotic figures" (left) and ``not mitotic figures" (right) on HPF subsections  from both an Apeiro scanner (top) and a Hamamatsu scanner (bottom).}
\label{fig:detections}
\end{figure}

See Fig. \ref{fig:detections} for sample detections.
\subsection{Metrics}
We used the F-measure (or $F_1$ score) score as the benchmark metric for our model. The F-measure score is defined as the harmonic mean of precision and recall:
\begin{equation}
F_1 = 2*\frac{{precision} * {recall}}{{precision} + {recall}},
\end{equation}
where precision and recall are defined as:
\begin{equation}
precision = \frac{TP}{TP+FP}, 
\end{equation}
\begin{equation}
recall = \frac{TP}{TP+FN},
\end{equation}
and $TP$ is the number of true positives, $FP$ is the number of false positives and $FN$ is the number of false negatives. \cite{mitosis-atypia-14} describe a true positive detection as a detection that is at most \SI{8}{\micro\metre} from the centroid of a ground truth mitosis.

\begin{figure}
\centering

\begin{tabular}{l|l|c|c|c}
\multicolumn{2}{c}{}&\multicolumn{2}{c}{Ground Truth}&\\
\cline{3-4}
\multicolumn{2}{c|}{}&Positive&Negative&\multicolumn{1}{c}{Total}\\
\cline{2-4}
\multirow{2}{*}{Prediction}& Positive & 72 & 31 & 103\\
\cline{2-4}
& Negative & 28 & 69 & 97\\
\cline{2-4}
\multicolumn{1}{c}{} & \multicolumn{1}{c}{Total} & \multicolumn{1}{c}{100} & \multicolumn{    1}{c}{100} & \multicolumn{1}{c}{200}\\
\end{tabular}

\begin{tabular}{l|l|c|c|c}
\multicolumn{2}{c}{}&\multicolumn{2}{c}{Ground Truth}&\\
\cline{3-4}
\multicolumn{2}{c|}{}&Positive&Negative&\multicolumn{1}{c}{Total}\\
\cline{2-4}
\multirow{2}{*}{Prediction}& Positive & 53 & 58 & 111\\
\cline{2-4}
& Negative & 47 & 42 & 89\\
\cline{2-4}
\multicolumn{1}{c}{} & \multicolumn{1}{c}{Total} & \multicolumn{1}{c}{100} & \multicolumn{    1}{c}{100} & \multicolumn{1}{c}{200}\\
\end{tabular}

\begin{tabular}{l|l|c|c|c}
\multicolumn{2}{c}{}&\multicolumn{2}{c}{Ground Truth}&\\
\cline{3-4}
\multicolumn{2}{c|}{}&Positive&Negative&\multicolumn{1}{c}{Total}\\
\cline{2-4}
\multirow{2}{*}{Prediction}& Positive & 96 & 5 & 101\\
\cline{2-4}
& Negative & $4$ & $95$ & $99$\\
\cline{2-4}
\multicolumn{1}{c}{} & \multicolumn{1}{c}{Total} & \multicolumn{1}{c}{100} & \multicolumn{    1}{c}{100} & \multicolumn{1}{c}{200}\\
\end{tabular}

\caption{Confusion matrices (from top to bottom): the standard Faster-RCNN, the Faster-RCNN with only $conv_4$ features, and our proposed model.}
\end{figure}

\begin{table}
\centering
 \begin{tabular}{||c c||} 
 \hline
 Method & $F_1$ score  \\ [0.5ex] 
 \hline\hline
 Faster-RCNN \citep{NIPS2015_5638} & 0.502  \\ 
 \hline
 Faster-RCNN w/ $conv_4$ & 0.709  \\ 
 \hline
  Proposed & \textbf{0.955}  \\ 
 \hline
\end{tabular}
\caption{Comparative performance of our proposed RCNN adaptation versus the standard Faster-RCNN model and a Faster-RCNN model with only $conv_4$ features.}
\label{fig:rcnns}
\end{table}

\begin{table}
\centering
 \begin{tabular}{||c c||} 
 \hline
 Reference & $F_1$ score  \\ [0.5ex] 
 \hline\hline
 \cite{10.1007/978-3-642-40763-5_51} & 0.782  \\ 
 \hline
 \cite{d00fb7c37eb248e1a28dc416fff2f8c3} & 0.611  \\
 \hline
 \cite{AAAI1611788} & 0.356  \\
 \hline
 \cite{7405343} & 0.433  \\
 \hline
 \cite{saha} &0.900  \\
 \hline
 Proposed & \textbf{0.955}  \\ 
 \hline
\end{tabular}

\caption{Comparative results involving recently published works and contest winners.}
\label{table:others}
\end{table}
\subsection{Comparative Results}

\subsubsection{Two-Stage Top-Down Cascade Multi-Scale Proposal Generation Results}

To display the efficacy of our custom two-stage top-down cascade multi-scale proposal generation method, we implemented a standard Faster-RCNN along with a Faster-RCNN only using $conv_4$ features and trained the models in a fashion identical to that of \cite{NIPS2015_5638}. Table \ref{fig:rcnns} shows the performance of our proposed model alongside the other 2 benchmark models. Our custom region proposal network improved upon the standard Faster-RCNN model by 90\% and the Faster-RCNN with only $conv_4$ features by 35\%.

\subsubsection{Comparison with Previously Proposed Methods}

Table \ref{table:others} shows the results of our model in comparison to recently published works and contest winners. Our approach was $6.22\%$ more accurate than the previous high $F_1$ score of 0.900 achieved by the model proposed by \cite{saha}. Our method outperformed all previously proposed approaches which utilized both handcrafted features and deep learning methodologies.

\subsubsection{Computation Time}

Most previous works do not document the time it takes for model evaluation. We found that, on average, our model took 0.5 seconds to process 1 HPF. \cite{saha} report that their model took 0.3 seconds per HPF. This increase in our model's forward propagation time can be attributed to the increased complexity of our model due to the need for 2 RPN's.

\section{Discussion}

\subsection{Significance}

Computerized extraction of mitotic counts from HPF's allows for a streamlined breast cancer prognosis pipeline. A biopsy taking 2 - 10 days \citep{bcancer} can be reduced significantly if manual mitotic figure counting was omitted from the current prognostic pipeline. Our contribution advances the state-of-the-art in computerized breast cancer prognosis, hopefully towards fully automated breast cancer prognosis in clinical practice. 

Outside the scope of computerized medical imaging, the detection of small-scale objects is a practice applicable to many problems. \cite{8019550} proposed that his model be used for company logo detection in images where logos make up small fractions of the image. Satellite images also have many small artifacts or landmarks scattered around large, high-resolution images. Detecting such small objects with great accuracy is made possible by our proposed model. The future applications are abundant and promising. 

\subsection{Limitations}

While models utilizing learned features may be more accurate than their counterparts relying on hand-crafted features, deep learning models of this scale have their limitations due to their computationally-exhaustive nature. Training a deep learning model like our proposed model requires GPU's in order to rapidly calculate extensive amounts of large-scale matrix operations along with large sets of varied, yet consistent data. Without multiple GPU's the time to complete training iterations can increase exponentially to infeasible values. Due to the advent of cloud computing power for deep learning models, we were able to address the issue of dealing with computationally taxing operations. However, there were more complications regarding the data which the model was trained, tested and evaluated on. While our dataset sources utilized the same set of tissue scanners (meaning relatively similar image characteristics), the histopathological images were from different labs using different staining protocols. Although we normalized the stain color across all images in our dataset, there would still be stain irregularities across our dataset which could result in increased false-positive or false-negative detections. Since our model was trained on data originally annotated by trained pathologists, our model was subject to human biases and error. The same variation of mitotic figures described in Section 1 also affects pathologists, meaning that misclassification are quite likely. Introducing a ``not mitotic figure" class to account for low-confidence pathologically-annotated mitotic figures was an attempt to solve this problem, but there is no guarantee that pathologists correctly identified all mitotic figures with relatively high confidence. Apart from the inconsistencies or irregularities within our dataset, the size of our dataset is a pertinent problem for deep learning models. Many large-scale models rely on massive datasets during the training process. We artificially increased the dataset size and variation through augmentation techniques, but our dataset was still quite small compared to the size of the datasets utilized by other works. For example, the VGG-16 model (our base feed-forward network) was trained on a dataset with upwards of a million samples of image data \citep{simonyan}.

\section{Conclusion}

We propose a novel variant of the Faster-RCNN architecture in order to detect mitotic figures in breast cancer histopathological images with great speed and accuracy. Our novel two-stage top down multi-scale region proposal generation process enables our model to detect small objects such as the mitotic figures. Our results reinforce the strength of our proposed model in comparison to previously proposed works. Our proposed model achieved an F-measure score of 0.955, the highest accuracy achieved to date. Our model utilizes purely learned features to detect mitotic figures, thus displaying the strength of learned features compared to the hand-crafted features used by other works.

\section*{Acknowledgements}

We gratefully acknowledge Professor Armando Fox (UC Berkeley) for assistance on formatting and writing the manuscript and both Smitha Rao (Stanford University) and Sanjay Krishnamurthy (Uber Technologies) for proofreading the manuscript.

\section*{References}
\bibliography{mybibfile}

\end{document}